%% file: main.tex
\documentclass[10pt,twocolumn,letterpaper]{article}

\usepackage[pagenumbers]{iccv} %

\input{definitions}

\input{preamble}

\definecolor{iccvblue}{rgb}{0.21,0.49,0.74}
\usepackage[pagebackref,breaklinks,colorlinks,allcolors=iccvblue]{hyperref}

\usepackage{multirow}
\usepackage{soul}
\usepackage{pythontex}
\usepackage{svg}
\usepackage{wrapfig}
\usepackage{algorithm}
\usepackage{algorithmic}

\def\paperID{1536}
\def\confName{ICCV}
\def\confYear{2025}

\title{\methodname: \\ Stochastic Rasterization for Sorting-Free 3D Gaussian Splatting}

\author{
  Shakiba Kheradmand*\textsuperscript{1,2},
  Delio Vicini*\textsuperscript{3},
  George Kopanas\textsuperscript{$\dagger$3,4},\\
  Dmitry Lagun\textsuperscript{1},
  Kwang Moo Yi\textsuperscript{2},
  Mark Matthews\textsuperscript{1},
  Andrea Tagliasacchi\textsuperscript{1,5,6}
  \\[1em]
  \textsuperscript{1}Google DeepMind, 
  \textsuperscript{2}University of British Columbia, 
  \textsuperscript{3}Google,
  \textsuperscript{4}Runway ML, \\
  \textsuperscript{5}Simon Fraser University,
  \textsuperscript{6}University of Toronto,\\
  *equal contribution, \textsuperscript{$\dagger$}work done at Google
}

\begin{document}

\input{figs/teaser/teaser}

\input{sec/0_abstract}

\input{sec/1_intro}
\input{sec/2_related_work}
\input{sec/3_method}

\input{sec/4_experiments}

\input{sec/5_conclusions}
{
    \small
    \bibliographystyle{ieeenat_fullname}

\input{main.bbl}
}
\input{sec/X_suppl}

\end{document}

%% file: definitions.tex
\newcommand{\methodname}{StochasticSplats\xspace}

\newcommand{\mipnerfdataset}{MipNerf360\xspace}
\newcommand{\stopthepop}{StopThePop\xspace}

\newcommand{\loss}{\mathcal{L}}

\newcommand{\mean}{\mu}
\newcommand{\erf}{\mathrm{erf}}
\newcommand{\covariance}{\boldsymbol{\Sigma}}
\newcommand{\opacity}{\alpha}
\newcommand{\sphcoef}{\mathbf{c}}
\newcommand{\renderweight}{\alpha_i \prod_{z_k < z_i} (1 - \alpha_k)}
\newcommand{\param}{\boldsymbol{\theta}}

\newcommand{\image}{\mathbf{I}}

\newcommand{\model}{\mathcal{G}}
\newcommand{\element}{g}

\newcommand{\origin}{\mathbf{o}}
\newcommand{\worldpos}{x}
\newcommand{\thresh}{\tau}

\newcommand{\vx}{\mathbf{x}}
\newcommand{\vd}{\mathbf{d}}
\newcommand{\vo}{\mathbf{o}}
\newcommand*\diff{\mathop{}\!\mathrm{d}} %

%% file: preamble.tex
\usepackage{bbm}
\usepackage{amstext}
\usepackage{amsmath}
\usepackage{xcolor}
\usepackage[format=plain,labelformat=simple,labelsep=period,font=small,compatibility=false]{caption}
\usepackage{xspace}
\usepackage{amssymb} 
\usepackage{bm}
\usepackage{mathtools}
\usepackage{comment}

\usepackage{float}

\newfloat{listing}{thp}{lop}
\floatname{listing}{Listing}

\usepackage{soul}
\setuldepth{foobar}

\def \customparskip {.2em}
\renewcommand{\paragraph}[1]{\vspace{\customparskip}\noindent\textbf{#1.}}

\definecolor{gold}{rgb}{0.7, 0.5, 0}
\definecolor{cvprblue}{rgb}{0.21,0.49,0.74}

\newcommand{\FLIP}{\protect\reflectbox{F}LIP\xspace}

%% file: figs/teaser/teaser.tex
\twocolumn[{%
\maketitle
\renewcommand\twocolumn[1][]{#1}%
\vspace{-3em}
\begin{center}
\includegraphics[width=\linewidth]{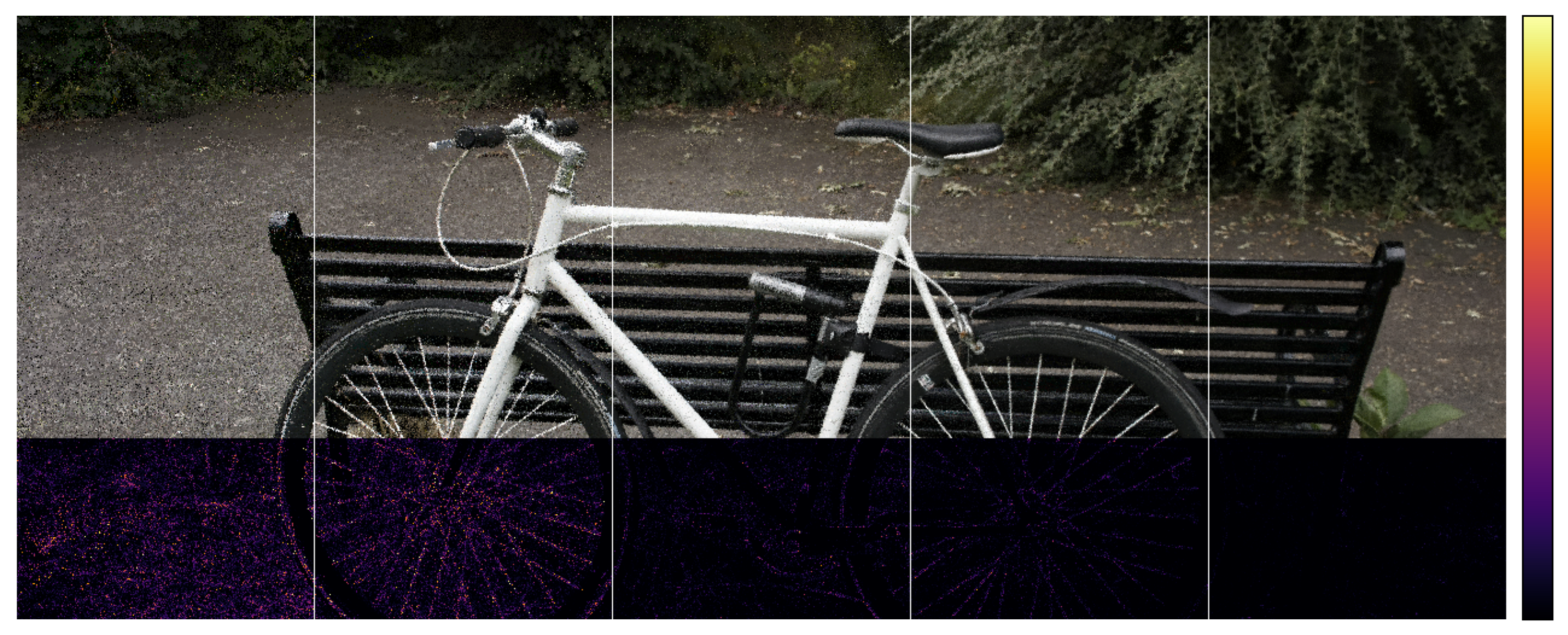}
\end{center}
\vspace{-1.5em}
\small
\setlength\tabcolsep{0pt} %
\begin{tabular}{*{5}{p{.194\linewidth}}}
    \centering 1~SPP, ~2.7~ms&
    \centering 2~SPP, ~2.9~ms&
    \centering 4~SPP, ~3.3~ms&
    \centering 8~SPP, ~5.6~ms&
    \centering 16~SPP, ~6.5~ms
\end{tabular}
\begin{tabular}{*{5}{p{.194\linewidth}}}
    \centering MSE: 0.0211 &
    \centering MSE: 0.0154 &
    \centering MSE: 0.0072 &
    \centering MSE: 0.0048 &
    \centering MSE: 0.0020
\end{tabular}
\vspace{-0.5em}
\captionof{figure}{\textbf{Teaser} --
We render a 3DGS radiance field using an unbiased Monte Carlo estimator of the volume rendering equation.
Unlike prior radiance field approaches, our method does not require sorted back-to-front rendering. 
This results in fast, portable, and pop-free rendering that is easy to implement.
The number of Monte Carlo samples per pixel allows a trade-off between interactivity and visual quality, similar to physically-based rendering.
From left to right, we increase the number of samples per pixel, which increases visual fidelity at a predictable increase in rendering cost.
We report the mean square error (MSE) and render time on an NVIDIA RTX 4090.
}
\vspace{2em}
\label{fig:teaser}
}]

%% file: sec/0_abstract.tex
\begin{abstract}
3D Gaussian splatting~(3DGS) is a popular radiance field method, with many application-specific extensions. Most variants rely on the same core algorithm: depth-sorting of Gaussian splats then rasterizing in primitive order. This ensures correct alpha compositing, but can cause rendering artifacts due to built-in approximations.
Moreover, for a fixed representation, sorted rendering offers little control over render cost and visual fidelity.
For example, and counter-intuitively, rendering a lower-resolution image is not necessarily faster.
In this work, we address the above limitations by combining 3D Gaussian splatting with stochastic rasterization. Concretely, we leverage an unbiased Monte Carlo estimator of the volume rendering equation.
This removes the need for sorting, and allows for accurate 3D blending of overlapping Gaussians.
The number of Monte Carlo samples further imbues 3DGS with a way to trade off computation time and quality.
We implement our method using OpenGL shaders, enabling efficient rendering on  modern GPU hardware.
At a reasonable visual quality, our method renders more than four times faster than sorted rasterization.
\end{abstract}

%% file: sec/1_intro.tex
\section{Introduction}
\label{sec:intro}
Recent advances in novel view synthesis have been driven by the success of neural radiance fields (NeRF) \cite{Mildenhall20eccv_nerf} and 3D Gaussian splatting (3DGS) \cite{Kerbl2023tdgs}. The former produced high-quality visuals at a low rendering speed, while the latter enabled efficient rendering with comparable visual quality.
3DGS represents a scene as a set of anisotropic 3D Gaussians that can be ``splatted''~\cite{Zwicker2001splatting} in rasterization pipelines, enabling real-time viewing and efficient training.

In 3DGS, 3D Gaussians are flattened onto camera-oriented billboards that not only prevent accurate volumetric intermixing~\cite{ever}, but must also be depthwise sorted for rasterization~\cite{zwicker2001ewa}.
This global sorting operation is computationally expensive and introduces popping artifacts~\cite{stopthepop}, as the sort order is \textit{view-dependent}: small camera movements can result in temporal discontinuities in the rendered images.
Further, there is no straightforward way to balance rendering cost vs. visual fidelity.
One might assume that reducing the rendering resolution would lead to faster rasterization... yet, we observe the opposite in our experiments.
Overall, this limits 3DGS's use on low-end hardware and in latency-critical applications, such as robotics and autonomous driving.
Finally, most implementation use custom software rasterizers or GPGPU frameworks such as CUDA.
Software rasterizers are notoriously hard to optimize, and often cannot match the performance of the hardware stack~\cite{Liu2019karas}.
Additionally, many GPGPU frameworks are vendor-specific, reducing portability.

Several previous works address popping artifacts, rendering performance, and portability, but typically improve only one, at the expense of the others.
Popping can be alleviated by per-pixel depth sorting, but this introduces extra computational demands. 
While this cost can be amortized using clever spatial biases~\cite{stopthepop}, per-pixel sorting cannot fully solve volumetric intermixing of 3D Gaussians.
Constant density ellipsoids allow analytical intermixing~\cite{ever}, but are slow to render because they replace sorting with expensive ray tracing.
Other methods~\cite{hahlbohm2024efficient} reduce the rendering cost and the popping artifacts by using a mix of per-pixel sorted lists and weighted order-independent transparency, but this comes at a cost in terms of portability since they need to implement complicated GPGPU kernels hindering cross-platform adoption.
An orthogonal route to reduce render cost is to reduce the number of Gaussians~\cite{kheradmand20243d}.
Finally, packages such as Splatapult~\cite{Thibault2024splatapult} implement an OpenGL-based rasterization pipeline, but do not outperform the original CUDA 3DGS implementation.

\input{figs/st_intro/st_intro}

We instead propose to render 3D Gaussians using stochastic transparency (ST)~\cite{enderton2010stochastic}; see~\Cref{fig:st_intro}.
Stochastic transparency replaces sorted alpha blending by an unbiased Monte Carlo estimator.
Practically, transparency is rendered by probabilistically selecting individual, opaque, samples that are averaged together with multi-sampling~\cite{Crow1977apc}.
Individual samples are rendered using order-independent Z-buffering, \textit{without} sorting.
We provide unbiased Monte-Carlo estimators of both the forward alpha blending and the gradient computation. The reverse-mode gradient estimator also permits sorting-free training and is largely agnostic to the primitive type and could be used for other semi-transparent representations~(e.g., triangle soups).
We also do an early analysis on proper volumetric intermixing of 3D Gaussians by extending the theory of stochastic transparency.
Critically, all of the above is naturally compatible with the hardware rasterization pipeline, allowing us a portable OpenGL implementation of our method.

At one sample per pixel (SPP), our OpenGL implementation renders up to $4\times$ faster than the original CUDA implementation, enabling latency-critical applications, such as open-vocabulary 3D object localization~\cite{kerr2023lerf}, which we demonstrate.
Higher SPP increase quality and render time, approaching the PSNR of alpha-blended 3DGS.
One key advantage is that downstream applications can \emph{dynamically} adjust sampling at rendering time, allowing them to trade-off latency against quality; see~\Cref{fig:teaser}.

%% file: figs/st_intro/st_intro.tex
\begin{figure}
\begin{center}
\includegraphics[width=\linewidth]{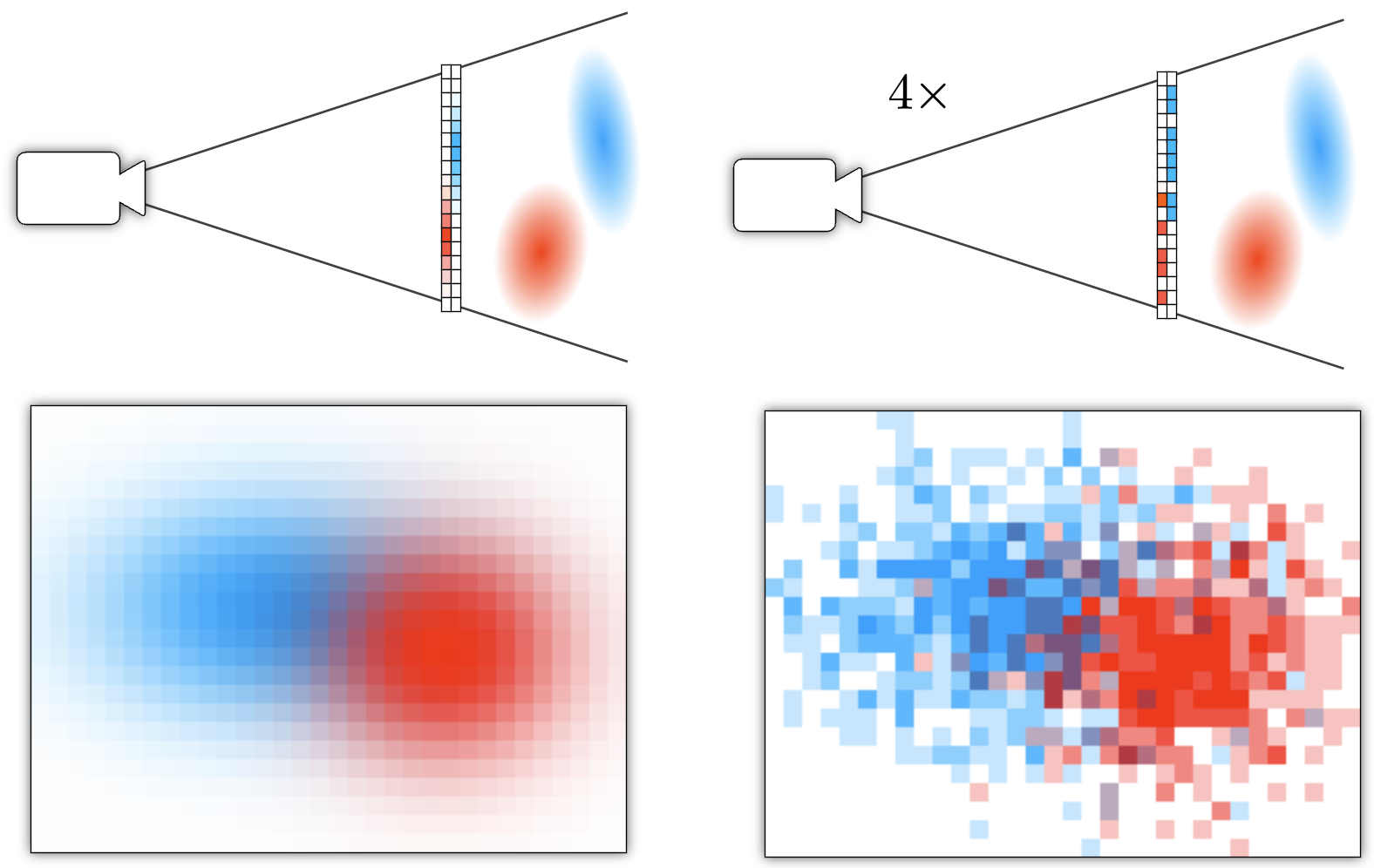}
\end{center}
\vspace{-1em}
\setlength\tabcolsep{0pt} %
\small
\vspace{-.4em}
\begin{tabular}{*{2}{p{.51\linewidth}}}
    \centering Alpha blending&
    \centering Stochastic transparency (4 SPP)
\end{tabular}
\vspace{-1.5em}
\caption{\textbf{Blending vs. stochastic} -- 
We visualize a 1D and 3D example of the different ways to compute transparency. 
Two foreground Gaussians are composited on a white background.
}
\vspace{-1em}
\label{fig:st_intro}
\end{figure}

%% file: sec/2_related_work.tex
\section{Related Work}
\label{sec:related_work}
\vspace{-\customparskip}
\paragraph{Order-independent transparency}
While current 3DGS methods rely on sorted alpha blending, there is a rich literature on order-independent transparency for rasterization-based real-time rendering.
Prior work can be classified into: (1)~stochastic methods~\cite{enderton2010stochastic, laine2011stratified, wyman2016stochastic}, that probabilistically discard fragments based on their opacity to create Monte Carlo estimators of the alpha blending equation and (2)~non-stochastic methods that avoid sorting by using per-pixel linked lists~\cite{yang2010real}, depth peeling~\cite{everitt2001interactive}, weighted blending~\cite{mcguire2013weighted}, moments~\cite{munstermann2018moment}, or deep learning~\cite{tsopouridis2024deep} to (approximately) composite transparent fragments.

Given the success of order-independent transparency in real-time graphics, it becomes desirable to incorporate these techniques with 3DGS. Recent pre-prints~\cite{hahlbohm2024efficient, hou2024sort} combine 3DGS with non-stochastic order-independent transparency, but stochastic variants remain unexplored. We bridge this gap by extensively studying stochastic transparency with 3DGS, both in terms of rendering and training, enabled by our novel stochastic gradient estimator.

\paragraph{Popping artifacts}
Real-time graphics eschew sorting primitives due to both the computational cost and the inexistence of a consistent global ordering of non-trivial primitives. In 3DGS this is overlooked, because most primitives have a relatively small spatial extent. This introduces artifacts often dubbed as ``popping'', where a small motion of the camera forces a change in the order of the primitives introducing a sharp discontinuity in the alpha blending equation. These artifacts are particularly noticeable in immersive contexts like virtual reality. \citet{stopthepop} address this using per-pixel sorting based on the maximal Gaussian response. While this eliminates popping artifacts, it introduces prohibitive computational costs, necessitating complex \emph{approximate} sorting through a multi-stage hierarchical renderer. EVER~\cite{ever} addresses popping artifacts by ray tracing constant density ellipsoids to compute the full volume rendering equation, rather than approximate alpha-blending. However, this approach is significantly slower than rasterization. In contrast, stochastic transparency uniquely eliminates popping by trading sorting for some noise in the final image. 

\paragraph{Fast 3DGS rendering and training}
While 3DGS~\cite{Kerbl2023tdgs} offered a significant speedup over NeRF-based methods, subsequent works have further improved performance. Numerous improvements have been made to the implementation of training and rendering routines~\cite{mallick2024taming, ye2024gsplatopensourcelibrarygaussian, hahlbohm2024efficient, gui2024balanced}. Further speedups can be made by reducing the number of Gaussians and compressing their attributes. This is covered in an extensive literature review~\cite{bagdasarian24073dgs}. Any approach that reduces the number of primitives readily complements our work and is likely to improve rendering performance. 

\paragraph{Differentiable Monte Carlo methods}
We introduce a novel Monte Carlo gradient estimator for stochastic transparency. Such differentiable Monte Carlo estimators are commonly used in physically-based inverse rendering \cite{gkioulekas2016evaluation,zhao2016, li18EdgeSampling, NimierDavidVicini2019Mitsuba2, zhang2019dtrt, Vicini2021PathReplay}. Various works propose differentiable Monte Carlo methods for neural radiance fields~\cite{verbin2023eclipse, mai2023neural} or surface reconstruction~\cite{hasselgren2022nvdiffrecmc}, and \citet{condor2024} differentiate physically-based Monte Carlo ray tracing of 3D Gaussian primitives. \citet{deliot2024transforming} applied stochastic simultaneous perturbation gradients to real-time rendering algorithms, including standard 3DGS. Beyond rendering, differentiable Monte Carlo methods find application in nuclear engineering~\cite{Mikhailov1966}, finance~\cite{Giles2007}, atmospheric physics~\cite{tregan2019convergence} and inverse partial differential equation problems~\cite{Yilmazer2024Solving}.

%% file: sec/3_method.tex
\section{Method}
\citet{Kerbl2023tdgs} represents a 3D scene in 3DGS as a collection of  anisotropic 3D Gaussians~$\model{=} \{ \element_i \} $.
Given a set of posed images $\{ \image_n \}$, we optimize the parameters of the 3D Gaussians $\model$ to photometrically reconstruct the scene.
Each $\element_i$ is parameterized as $\param {=} \{\mean,\covariance,\opacity,\sphcoef\}$, where $\mean$ is the mean, $\covariance$ is the covariance matrix, $\opacity$ is the opacity, and $\sphcoef$ is the view-dependent color modeled using spherical harmonics~\cite{Ramamoorthi2001sph}.
We start by reviewing the classical alpha-blending based rendering of 3DGS in \Cref{sec:blending}, and then introduce our stochastic transparency algorithm in \Cref{sec:stochastic}.

\subsection{Sorted alpha blending}
\label{sec:blending}
To calculate the radiance $C$ at a given pixel, we aggregate the radiances emitted by $L$ primitives that intersect the associated ray, 
blending them in depth ($z$) order~\cite{Porter1984cdi}: 
\begin{equation}    
    C = \sum_{i=1}^L c_i \renderweight
    \label{eq:ab}
\end{equation}
where $z_k {<} z_i$ denotes all primitives in front of primitive $i$.
While it can be shown that this expression derives from volume rendering~\cite{digest}, the product of terms in \cref{eq:ab} requires sorting the primitives along $z$, then blend radiances in either back-to-front or front-to-back order. As sorting is \textit{independent} of how many pixels one needs to render, we simply lose the ability to trade off accuracy with speed.

Sorting is done with respect to the $z$ coordinates of the Gaussian means $\mu_k$ remapped to the coordinate frame of the current camera, and the covariance is projected via a linear approximation of the non-linear projection process to the camera's view point, as proposed by~\citet{zwicker2001ewa}.
This effectively treats each Gaussian as a \textit{billboard}, flattening it parallel to the camera's imaging plane.

\subsection{Stochastic transparency}
\label{sec:stochastic}
Stochastic transparency~(ST)~\cite{enderton2010stochastic} avoids sorting by stochastically estimating alpha blending in~\cref{eq:ab}.
Avoiding sorting when rendering transparent elements is critical, because
\begin{enumerate*}[label=(\roman*)]
\item primitives often do not have a global unique sorting order across all pixels,
\item sorting can be computationally expensive, and 
\item in the context of volume rendering the 3D extent of primitives  influences the intermixing of colors.
\end{enumerate*}

Stochastic transparency \textit{approximates} transparency by randomly assigning a number of fully opaque samples per pixel proportional to a fragment's opacity, and averaging these samples to produce the final pixel color \textit{without} sorting.\footnote{Note that as we do not use multi-sampling anti-aliasing, making fragments synonymous with pixels.}
In our context, the opacity of a fragment is proportional to the opacity of the Gaussian $\alpha$, and to its distance from the mean $\mu$, as measured by the \textit{projected} covariance matrix~$\Sigma$. Stochastic transparency can also be interpreted as a randomized extension of classical \textit{screen-door transparency}~\cite[Chapter 16]{foley1996book}, where, instead of a randomized selection, a fixed pattern is employed to discard fragments.

\paragraph{Simple example}
As a simple example, consider a \textit{single} 3D Gaussian $g$ covering pixel $p$ with opacity $\alpha{=}0.5$; in this case, if we sample $x {\sim} \mathcal{U}(0,1)$ and accept $g$ only when $x{>}\alpha$, approximately $50\%$ of the pixel's samples will randomly select this Gaussian. We next formalize this idea to generalize to rendering an arbitrary number of Gaussians.

\paragraph{Formalization}
Formally, stochastic transparency is a Monte Carlo estimator of \cref{eq:ab}.
It estimates the radiance~$C$ by randomly sampling a  summand~$i$ (\ie, the $i$-th Gaussian):
\begin{align}    
    C \approx \frac{1}{P(i)} \: c_i \renderweight, 
    \label{eq:montecarlo}
\end{align}
where~$P(i)$ is the probability mass function of the random index~$i$.
This estimator is unbiased, as long as $P(i){>}0$ for any non-zero summand.
The standard approach is to use:
\begin{align}
P(i) = \renderweight.
\label{eq:alpha_pmf}
\end{align}
The division by this probability mass function cancels out part of the sample weight, leading to a simple estimator:
\begin{equation}
C \approx c_i \quad\text{for}\quad i \sim P.
\end{equation}
Thus, a pixel has the probability $P(C{=}c_i){=}P(i)$ to pick color $c_i$. In practice, we average multiple independent samples per pixel to reduce estimator noise.

\input{figs/example2}

\paragraph{Pseudocode}
Listing~\ref{listing:st} provides pseudocode for the above estimator. 
By testing that the depth of the current sample is smaller than the previously selected one, we sample exactly according to $P(i)$~\cite{enderton2010stochastic}. In the fragment shader, this is implemented by using standard depth testing and calling \texttt{discard} on samples where $u {\geq} \alpha_i$. This only requires access to the current Gaussian's opacity and works for any primitive ordering.
\Cref{fig:st_intro} further shows a toy example of using this estimator on two Gaussians.

\subsection{Differentiable stochastic transparency}
\label{sec:st-gradients}
Applying stochastic transparency to inverse rendering problems requires computing the gradient of the photometric loss $\loss$ with respect to the primitive parameters $\param$.
Primarily, we require gradients with respect to color ${\partial \loss}/{\partial c}$, and gradients with respect to opacity ${\partial \loss}/{\partial \alpha}$. 

\paragraph{Gradient estimator} We take inspiration from the differentiable rendering literature~\cite{Vicini2021PathReplay}, and use a \emph{detached} gradient estimator~\cite{Zeltner2021MonteCarlo, Vicini2021PathReplay} to correctly differentiate the binary sampling decision in Listing~\ref{listing:st}. We derive this estimator by applying the gradient operator to the alpha blending equation:
\begin{equation}    
    \nabla_{\param} C = \sum_{i=1}^L \nabla_{\param} \left[ c_i \renderweight \right].
    \label{eq:ab_grad}
\end{equation}%
We then form an unbiased Monte Carlo gradient estimator by again sampling a summand according to $P(i)$. Crucially, we explicitly do not differentiate $P(i)$ itself and obtain:
\begin{equation}
\nabla_{\param} C \approx \frac{1}{P(i)} \nabla_{\param} \left[ c_i \renderweight \right ].
\end{equation}
Substituting the definition of $P(i)$ from \cref{eq:alpha_pmf} and differentiating $\loss$ with respect to $c$ and $\alpha$, we get:
\begin{align}
    \frac{\partial \loss}{\partial c_i} &= \frac{\partial \loss}{\partial C} &
    \frac{\partial \loss}{\partial c_{z_k < z_i}} &= 0
    \label{eq:gradients1}
    \\
    \frac{\partial \loss}{\partial \alpha_i} &= \frac{\partial \loss}{\partial C} \frac{c_i}{\alpha_i} &
    \frac{\partial \loss}{\partial \alpha_{z_k < z_i}} &= \frac{\partial \loss}{\partial C} \frac{-c_i}{1 - \alpha_{z_k < z_i}},
    \label{eq:gradients2}
\end{align}
where subscript $i$ denotes the sampled primitive, and  ${\partial \loss}/{\partial C}$ is the photometric loss gradient, dependent upon the loss formulation~(L1, SSIM, ...).
Note how in \cref{eq:gradients1} only the $i$-th Gaussian  receives gradients with respect to its \textit{color}, while the others have a zero gradient.
Meanwhile, \cref{eq:gradients2} shows how the $i$-th Gaussian, and ones closer to the camera receive a gradient with respect to their \textit{opacity}.
The Gaussians behind the sampled one \textit{do not receive any gradient}.
Finally, our derivation is not specific to 3DGS, and applies to \emph{any} representation rendered with stochastic transparency.

\input{figs/gradients/item}

\paragraph{Decorrelated loss gradient} 
A well-known challenge for differentiable Monte Carlo methods is that the derivative of the loss function is a product of two terms that are corrupted by noise (\ie, they are both random variables):
\begin{align}
\frac{\partial \loss}{\partial \theta} = \frac{\partial \loss}{\partial C} \frac{\partial C}{\partial \theta}.
\end{align}
where $\theta$ are the geometry parameters.
If the noise in both terms is \emph{correlated}, we get 
biased gradients, which harm optimization.
Following~\cite{gkioulekas2016evaluation, azinovic2019}, we use different random samples to evaluate the two factors above. 
If $\loss$ is the L2 loss, the decorrelated evaluation results in unbiased gradients. For other loss functions, it at least reduces bias. Unbiased gradient estimation for arbitrary loss functions applied to a Monte Carlo estimator is an open problem.

\paragraph{Practical algorithm} The presence of $c_i$ in~\cref{eq:gradients2} implies that we need access to the result of the forward pass to compute gradients. We therefore follow \emph{path replay backpropagation}~\cite{Vicini2021PathReplay} and compute gradients in three passes:
\begin{enumerate}
    \item In a first pass, we render an image and evaluate ${\partial \loss}/
    {\partial C}$.
    \item We then render a decorrelated image using a different random seed, giving us $c_i$ 
    \item Finally, we \emph{replay} the second pass by re-rendering using the \emph{same} random seed and using the stored $c_i$ to compute parameter gradients.
\end{enumerate}

\paragraph{Validation}
We validate our estimator by comparing to the gradients of the sorted alpha blending implementation. For easier visual interpretation, we use our reverse-mode gradient implementation to compute a \emph{gradient image}. We do this by backpropagating an L1 loss separately for each pixel, and visualizing the sum of the x-component of the positional gradients of all Gaussians affecting a given pixel. As shown in \Cref{fig:gradients}, our method produces noisy gradients that closely match those of alpha blending in structure.

\input{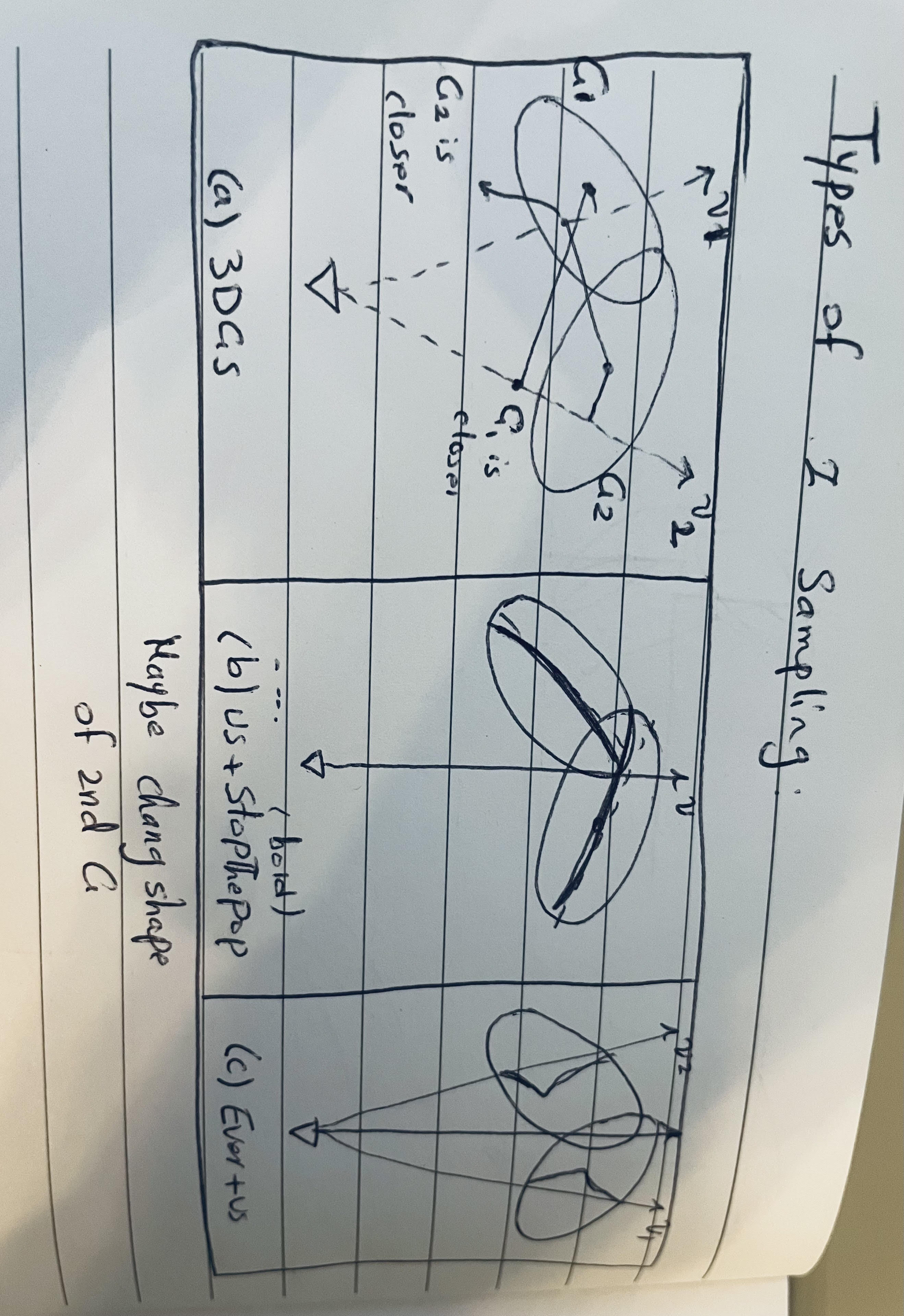}

\subsection{Removing popping artifacts}
\label{sec:popping_artifacts}
A naive implementation of stochastic transparency re-uses the billboard geometry of 3DGS, and consequently, to compute the depths~$z_i$, the means of the Gaussians.
As we will see in~\Cref{sec:forward_time}, at low sample count this provides high rendering efficiency, as we do not require a global sort.
However, popping artifacts still exist, as the computation of~$z_i$ is analogous to the one in 3DGS, small changes in the camera rotation can therefore change the order of primitives, and cause visible \textit{video} artifacts; see~\Cref{fig:z_methods}.

\paragraph{The stop-the-pop way}
\citet{stopthepop} overcome the popping problem by computing the per-fragment (vs. per-Gaussian) depth.
Their core idea is to compute the location of the \textit{peak} of the Gaussian density along the ray.
When coupled with a (computationally expensive) per-pixel sorting of Gaussians, this can provide a pop-free rendering.
This can be interpreted as finding ``maximum density surface'' of the Gaussian, visualized in 2D in~\Cref{fig:z_methods}, and using it to define fragment depths.
However, per-fragment updates of depth are not particularly efficient, as early-discard mechanisms implemented in hardware~\cite{khronos} can only execute for fragments whose depth has \textit{not} been modified by a shader.
While in our setting we can forego sorting, the inability to early-discard fragments still affects us, which lead us to propose a different approach, described next.

\paragraph{Our simplified way}
We propose to avoid this inefficiency by foregoing per-fragment depth changes.
Instead, we orient our billboards to linearly \textit{approximate} the maximum density surface.
If the ray associated with a pixel has origin $\origin$, the plane can be defined as:
\begin{equation}
\mathbf{n}^\top \bigl(\mathbf{x} - \boldsymbol{\mu}\bigr) \;=\; 0,
\quad
\text{where}
\quad
\mathbf{n} 
\;=\; 
\Sigma^{-1}\bigl(\boldsymbol{\mu} - \mathbf{o}\bigr)
\end{equation}
By approximating this surface with a plane, we can easily use the interpolation kernels within the hardware rasterizer to compute the splat's depth.
This approach adds no extra time to the render process and gives us popping-free rendering with a similar PSNR to~\cite{stopthepop} at higher sample counts.

\input{figs/intermix/item}
\paragraph{Fully volumetric intermixing}
If maximum rendering fidelity is desired and the cost of per-fragment depth modification is acceptable, stochastic transparency can \emph{naturally} support full volumetric intermixing. The key insight is that this can be achieved by replacing the fixed depth of a Gaussian by a sampled \emph{free-flight distance}. For this, we first express the total contribution of a single Gaussian using the \emph{Beer-Lambert law}:
$
    \int_0^{\infty} \sigma_i(\vx_t) \exp\left(-\int_0^t \sigma_i(\vx_s) \diff s\right) c_i \diff t,
$
where the opacity of a Gaussian is reparametrized using a spatially-varying \emph{extinction} function $\sigma_i$ and $\vx_a {\coloneqq} \vo {+} a {\cdot} \vd$ for a ray $(\vo, \vd)$. 
We follow \cite{condor2024} and sample a free-flight distance $t$ proportional to $p(t) = \sigma_i(\vx_t) \exp(-\int_0^t \sigma_i(\vx_s) \diff s)$. 
To handle overlapping Gaussians, we adapt \emph{decomposition tracking}~\cite{Novak2014RRT}:
If two Gaussians overlap, we sample a free-flight distance according to each and take the minimum. This accurately resolves overlap of an arbitrary number of Gaussians without additional data structures, sorting, or ray tracing, and readily fits into the stochastic transparency algorithm. We simply need to replace the computation of $z_i$ by a probabilistic sampling $z_i \sim p(t)$. We show an example rendering of two overlapping Gaussians in \cref{fig:intermix}. See the supplemental for the full free-flight sampling routine.

%% file: figs/example2.tex
\begin{listing}[t]
\begin{center}
\fbox{\begin{minipage}{.95\linewidth}
\begin{algorithmic}
\STATE $z \leftarrow \infty$ , C $\leftarrow c_{background}$
\FOR{each Gaussian $i$ in \ul{unsorted} order}
    \STATE Compute $\alpha_i$ and depth $z_i$
    \STATE Sample $u \sim \mathcal{U}(0,1)$
    \IF{$u < \alpha_i$ \AND $z_i < z$}
        \STATE C $\leftarrow$ $c_i$
        \STATE $z \leftarrow z_i$ 
    \ENDIF
\ENDFOR
\end{algorithmic}
\end{minipage}}
\end{center}
\vspace{-1.5em}
\caption{
Pseudocode for stochastic transparency.
}
\label{listing:st}
\end{listing}

%% file: figs/gradients/item.tex
\begin{figure}
  \centering
  \includegraphics[width=\linewidth]{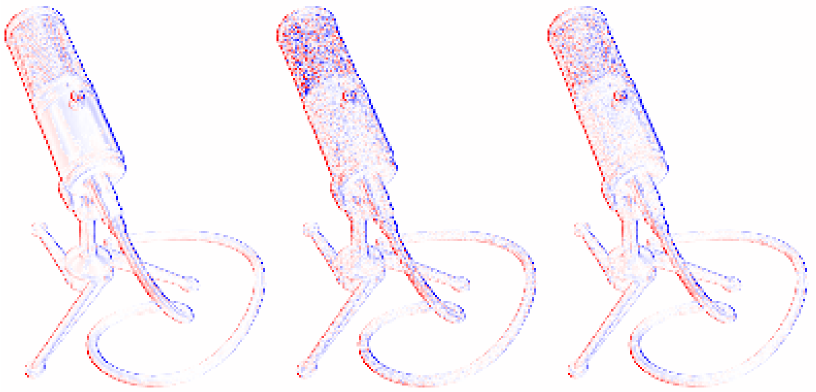}
\vspace{-1.5em}
\setlength\tabcolsep{0pt} %
\small
\begin{tabular}{*{3}{p{.33\linewidth}}}
    \centering Alpha blending&
    \centering Ours (128 SPP)&
    \centering Ours (512 SPP)
\end{tabular}  
\vspace{.1em}
\caption{\textbf{Gradient validation} -- We compare alpha blending gradients  to our stochastic estimator with 128 and 512 SPP. Red and blue colors encode positive and negative values, respectively. Our estimator accurately approximates the alpha blending gradients.}
\label{fig:gradients}
\vspace{-1em}
\end{figure}

%% file: figs/z_methods/z_methods.tex
\begin{figure}
\centering
\includegraphics[width=\linewidth]{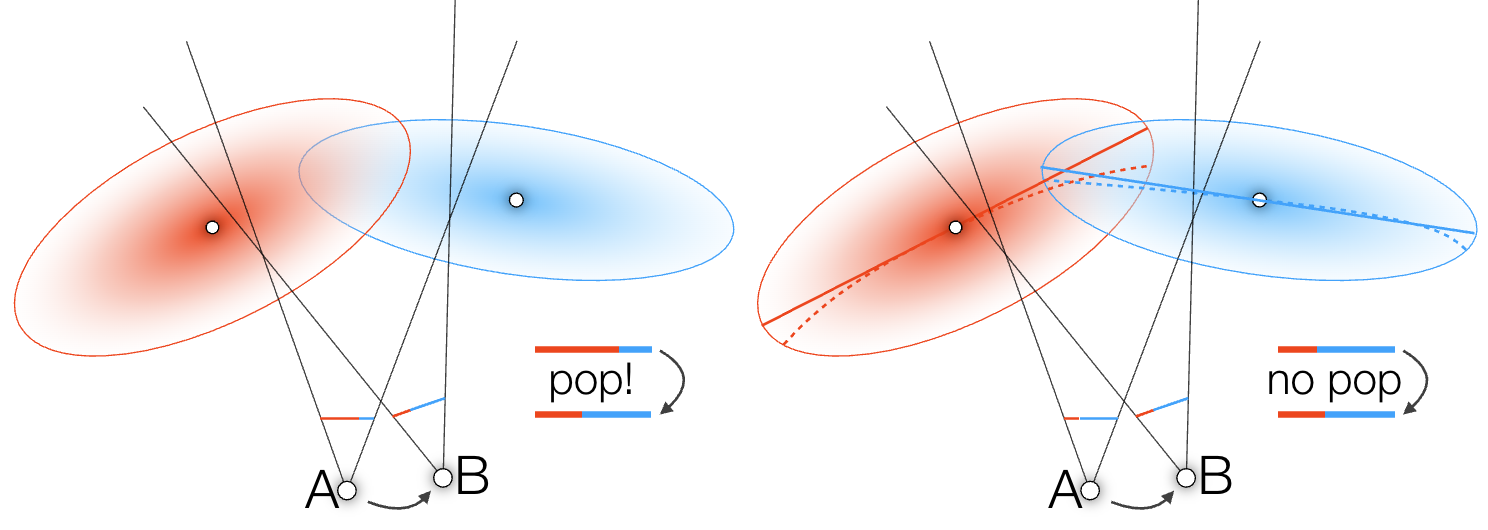}
\vspace{.5em}
\small
\setlength\tabcolsep{0pt} %
\begin{tabular}{*{2}{p{.499\linewidth}}}
    \centering 3DGS &
    \centering \stopthepop~\cite{stopthepop} vs. ours
\end{tabular}
\vspace{-1.8em}
\caption{
\textbf{Popping in 3DGS --}
(left) As sorting in 3DGS is done with respect to the $z$ distance of the Gaussian mean from the camera, a small camera rotation can cause visible ``popping'' artifacts~(\ie,~sudden pixel color changes). (right) \stopthepop~\cite{stopthepop} corrects for this behavior by associating a surface with each Gaussian, and determining the depth $z$ per-fragment (dashed line), rather than per-Gaussian. Our solution leads to visually comparable results, but approximates this surface linearly (solid line), so that per-fragment depth $z$ can be computed efficiently in hardware.
}
\label{fig:z_methods}
\vspace{-1em}
\end{figure}

%% file: figs/intermix/item.tex
\begin{figure}
\begin{center}
\includegraphics[width=0.99\linewidth]{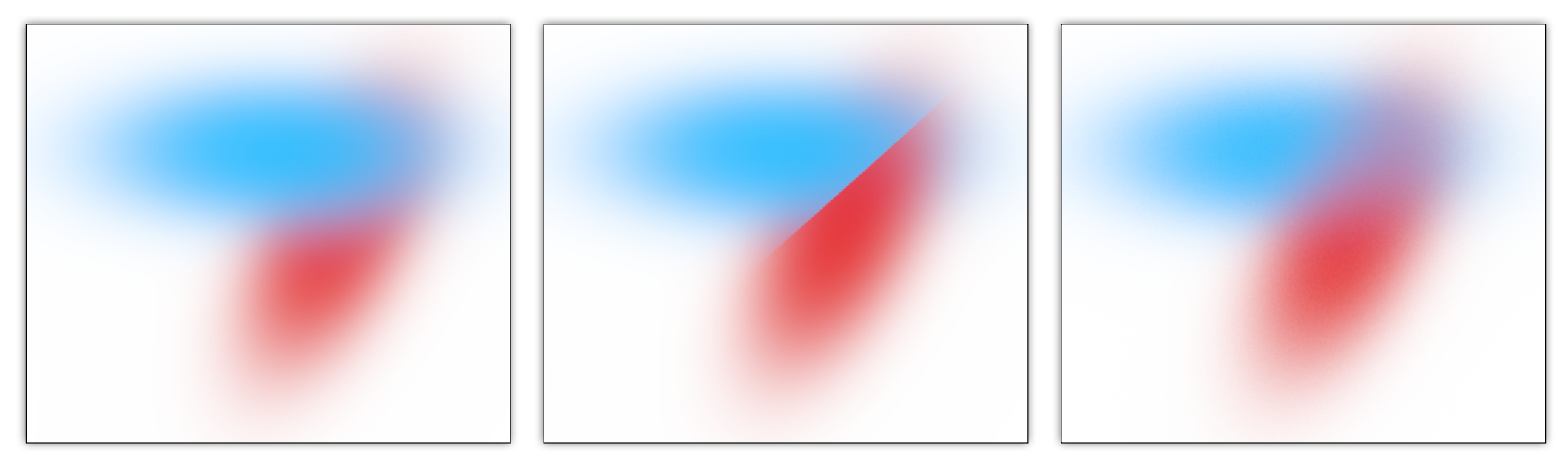}
\end{center}
\vspace{-1.3em}
\setlength\tabcolsep{0pt} %
\small
\begin{tabular}{*{3}{p{.33\linewidth}}}
    \centering Alpha blending&
    \centering Pop-free&
    \centering Our full intermixing
\end{tabular}  
\vspace{-1em}
    \caption{
    {\bf Fully volumetric intermixing} -- While alpha blending (left) suffers from popping and ``pop-free'' variants (middle) show discontinuities, our full volumetric intermixing (right) accurately composes overlapping Gaussians.
    }
    \vspace{-1em}
    \label{fig:intermix}
\end{figure}

%% file: sec/4_experiments.tex
\section{Experiments}
\label{sec:experiments}

We evaluate rendering performance on the real-world scenes from the \mipnerfdataset~\cite{Barron2022mipnerf360} dataset. 
For demonstrating semantic object localization with single-sample rendering we use the LERF dataset \cite{kerr2023lerf} -- it consists of complex in-the-wild scenes, designed for 3D object localization.

\paragraph{Baselines} 
For rendering baselines, we compare the performance of our method against the original 3DGS implementation~\cite{Kerbl2023tdgs} that uses CUDA, as well as Splatapult~\cite{Thibault2024splatapult}, an equivalent OpenGL implementation. We also compare with  StopThePop~\cite{stopthepop}, a pop-free method for 3DGS rendering.

\paragraph{Metrics} 
We compute the standard metrics of peak signal-to-noise ratio~(PSNR), structural similarity index metric~(SSIM)~\cite{wang2004ssim}, and learned perceptual image patch similarity~(LPIPS)~\cite{zhang2018perceptual}. 
We report the average per-frame processing time for all methods on the \mipnerfdataset scenes.

\paragraph{Implementation details}
To demonstrate portability, we implement our method in \textit{both} CUDA and OpenGL.
Our CUDA implementation builds on the existing 3DGS framework by \citet{Kerbl2023tdgs} and our OpenGL implementation builds upon Splatapult~\cite{Thibault2024splatapult}.
We evaluate on a Windows laptop equipped with an NVIDIA Quadro T1000 Max-Q GPU, and Linux desktops with an NVIDIA RTX 3090 and RTX 4090.
\quad
Rather than training from scratch, we fine-tune a trained 3DGS scene with the CUDA implementation of our method for 1000 iterations.
As we are fine-tuning, we do not apply densification or pruning, and we optimize positions using a learning rate of $5{\times}10^{-5}$ with 128 samples.
We keep all other hyperparameters the same. On average, the finetuning process requires approximately 14 minutes on an NVIDIA RTX 3090 GPU for scenes from~\mipnerfdataset.

\subsection{Rendering quality}
\label{sec:render_eval}

\input{figs/popfree/pop}
\input{figs/qualitative/item}

\vspace{-\customparskip}
\paragraph{Qualitative results -- \cref{fig:qualitative,fig:pop}}
We show two scenes~(\emph{Room} and \emph{Kitchen}) in \Cref{fig:qualitative} from the \mipnerfdataset dataset.
As shown, our method allows a trade-off between quality and rendering time.
\Cref{fig:pop} further highlights that popping is handled properly.
Note that, due to the sorting-free nature of our method, any popping free formulations can be easily integrated into our method.

\input{tables/rendering_eval}

\paragraph{Quantitative results -- \cref{tab:forward-only-psnr-avg}}
We report the PSNR values of our fine-tuned scenes rendered with varying numbers of samples, alongside comparisons to 3DGS and \stopthepop. 
We use our pop compensation from~\Cref{sec:popping_artifacts}. 
By increasing the number of samples per pixel, our method approaches the quality of alpha blending.
While the Monte Carlo noise affects metric, i.i.d. noise is visually often less distracting than structured artifacts.
Additionally, as we show in \Cref{par:taa_warping} and \Cref{sec:segmentation_eval}, this can be mitigated via \textit{temporal anti-aliasing}, and even without any mitigation, does not harm downstream perception tasks.

\subsection{Computation time}
\label{sec:timings_eval}

\input{tables/forward_time_eval}
\input{tables/training_eval}

\paragraph{Forward pass timings  -- \cref{tab:forward-only-fps-avg}}
\label{sec:forward_time}
We compare the forward rendering time of our method implemented in OpenGL, against 3DGS on both CUDA and OpenGL, as well as \stopthepop~\cite{stopthepop} on CUDA.
We report timings with a varying number of samples per pixel, exercising our method's ability to trade off performance and quality.
\quad
With a single sample, our method achieves $2\times$ to $4\times$ lower latency compared to alpha-blended 3DGS, depending on the hardware; see~\Cref{tab:forward-only-fps-avg}. For higher sample counts, we employ supersampling: render at a higher resolution and then downsample to the target display resolution. 

Our method is unique at being fast, portable and pop-free at the same time.
The portability of our method is only comparable with the 3DGS-OpenGL variant, that has popping artifacts and is relatively slow.
\stopthepop cannot be easily ported to standard graphics frameworks, and 3DGS-CUDA slows down significantly when ported to OpenGL.
This is because the CUDA radix sort routines utilize platform specific optimizations~\cite{polok2014fast}.

\input{figs/downsample/item}
\paragraph{Backward pass timings -- \cref{tab:backwards_eval}}
We further report the computation time for a single backward pass, this time with the CUDA implementation for all methods, as training typically is done on CUDA devices.

\subsection{Lowering resolution does not enhance speed}
\label{sec:naive_eval}

Our method inherently offers a performance vs. quality trade-off. 
Attempting to naively integrate such a trade-off into 3DGS, such as reducing rendering resolution, does not work well.
Simply lowering the rendering resolution does not speed up rasterization -- the same number of Gaussians must be processed, and the view frustum for each pixel increases.
In short, more Gaussians contribute to each pixel, increasing the computation time per pixel.
\Cref{fig:downsample} reports rendering times for our method (OpenGL) and alpha blending (AB-CUDA) with various tile sizes.
While AB render time decreases to some degree, it starts to increase after a point, as the Gaussian count per-tile increases.
No configuration can reach the rendering speed of our method (1 SPP).

\input{figs/taa/item}
\subsection{Making the most out of a single sample}
\label{sec:single_sample_eval}
We now practically demonstrate that even a single-sample rendering strategy can be useful.

\paragraph{Temporal anti-aliasing -- 
\label{par:taa_warping}
\cref{fig:taa_warping}}
In interactive applications, multiple frames are rendered in rapid succession.
A common strategy to further reduce variance is to average pixel values across frames by reusing information from previous renderings~\cite{Scherzer2012tcm}.
We employ temporal anti-aliasing~(TAA) to not only mitigate aliasing artifacts, but also accumulate samples over time, achieving an effectively higher sample count at minimal cost. 
Frame-to-frame consistency is maintained by \emph{warping} previously rendered frames according to per-pixel depth whenever the camera moves. As shown in \Cref{fig:taa_warping}, this greatly reduces rendering noise, while maintaining low latency.
We provide implementation details in the appendix, and our supplemental video demonstrates TAA applied to 1-SPP frames.

\input{figs/segmentation/seg}

\input{tables/segmentation_eval}

\paragraph{Semantic localization -- \cref{fig:segmentation} and \cref{tab:localization_eval}}
\label{sec:segmentation_eval}
We further show that while the single-sample render has low PSNR, it is still useful.
Specifically, we use LangSplat~\cite{qin2024langsplat} and replace their alpha blending renderer with our method, and follow their pipeline to localize objects.
As shown in \Cref{fig:segmentation}, while our renders are noisy, it still properly renders semantics such that the localizer can correctly identify its location.
There is a moderate reduction in accuracy~(\cref{tab:localization_eval}), but we achieve $4\times$ faster rendering speed.

%% file: figs/popfree/pop.tex
\begin{figure}
    \centering
    \includegraphics[width=\linewidth]{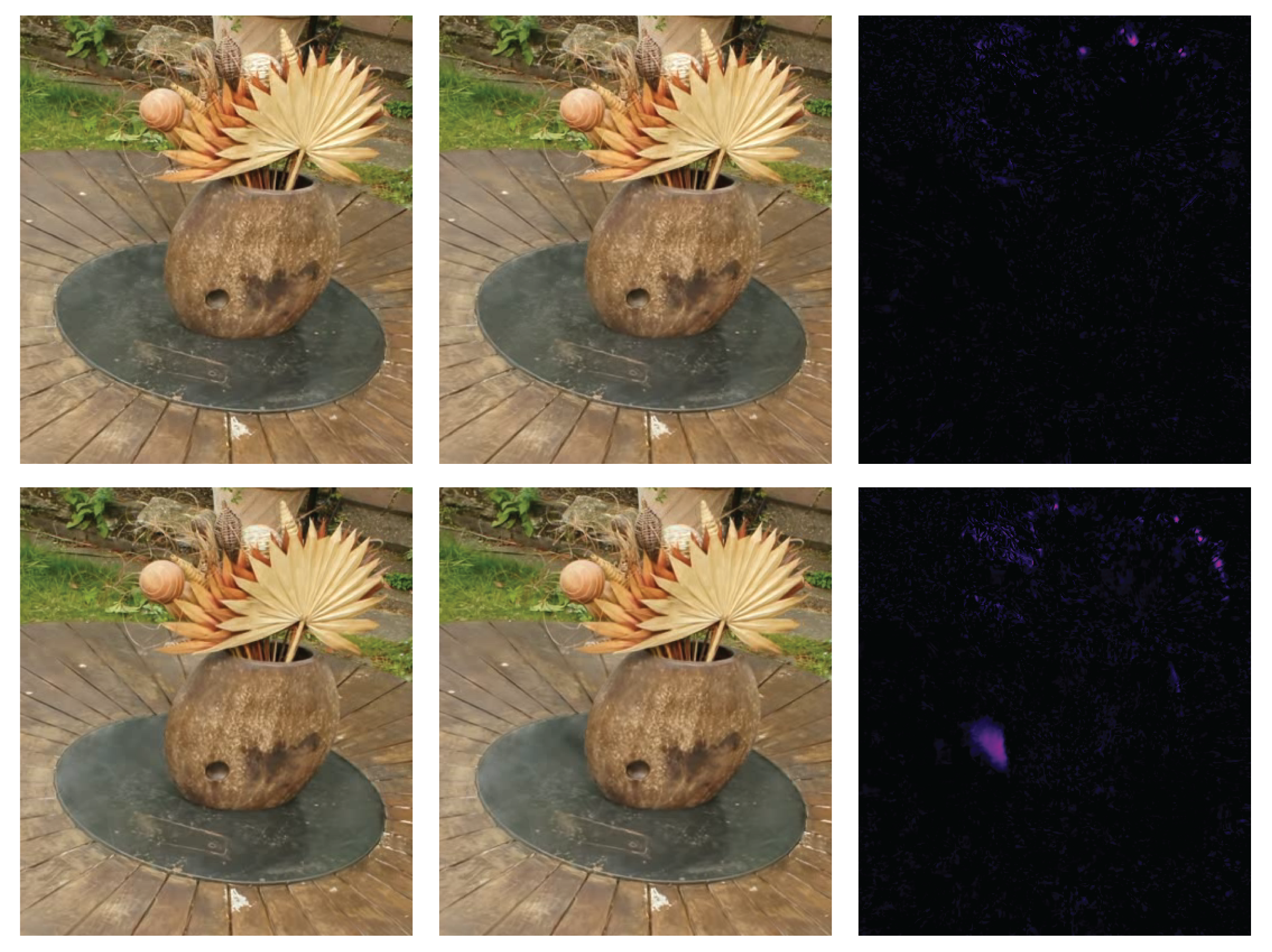}
    \setlength\tabcolsep{0pt} %
\small
\begin{center}
\vspace{-1.2em}
\begin{tabular}{*{3}{p{.3\linewidth}}}
    \centering Frame $t$&
    \centering Frame $t{+}1$&
    \centering \FLIP error
\end{tabular}
\end{center}
    \vspace{-1.7em}
    \caption{
    {\bf Popping artifact} -- 
   The top row displays two consecutive frames, fine-tuned and rendered using our \methodname, along with their corresponding \FLIP error map \cite{Andersson2020}. 
   The bottom row presents the same frames processed with standard 3DGS. 
   The foreground error indicates popping while the background errors are due to the camera motion.
    }
    \vspace{-1.2em}
    \label{fig:pop}
\end{figure}

%% file: figs/qualitative/item.tex
\begin{figure*}
\centering
\includegraphics[width=\linewidth]{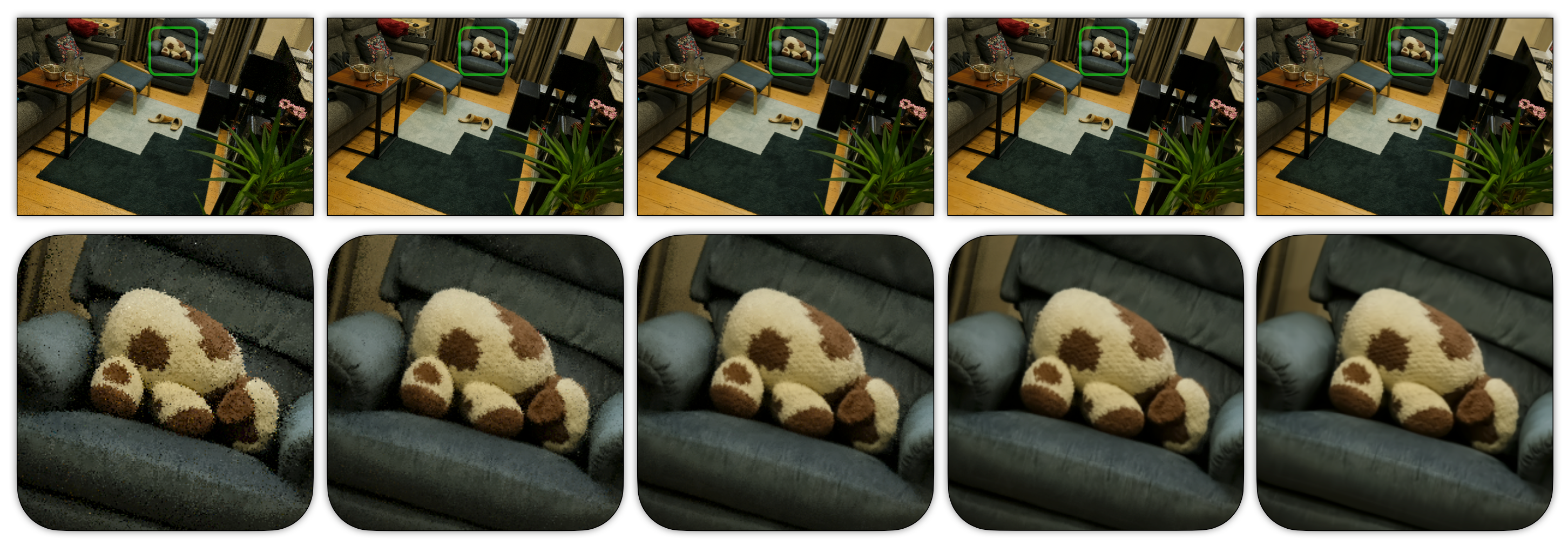}
\includegraphics[width=\linewidth]{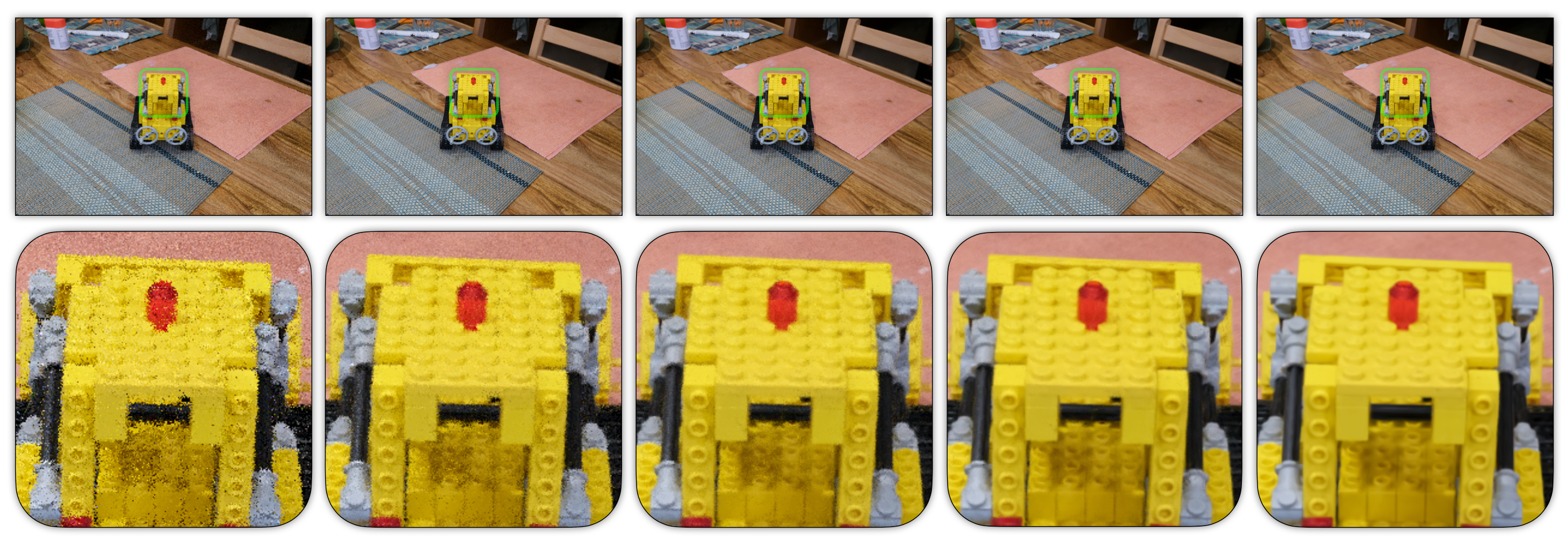}
\\[-.5em]
\setlength\tabcolsep{0pt} %
\small
\begin{tabular}{*{5}{p{.196\linewidth}}}
    \centering 1~SPP&
    \centering 4~SPP&
    \centering 16~SPP&
    \centering 128~SPP&
    \centering Alpha blending
\end{tabular}  
\vspace{-1em}
\caption{\textbf{Qualitative results -- } for the \textit{Room} and \textit{Kitchen} scenes from the \mipnerfdataset dataset are shown in \cref{fig:qualitative}. We compare \methodname using 1, 4, and 16 samples per pixel against the standard alpha blending approach. Even with just 1 sample per pixel, our method produces reasonable renderings, and with 16 samples, the results closely match the quality of alpha blending. 
}
\vspace{-1em}
\label{fig:qualitative}
\end{figure*}

%% file: tables/rendering_eval.tex
\begin{table}
    \centering
    \resizebox{\linewidth}{!}{
    \setlength{\tabcolsep}{8pt}
     \begin{tabular}{@{}lccccccccccc@{}} \toprule
     &  &  &\multicolumn{6}{c}{Ours (SPP)} \\ 
     & 3DGS & \stopthepop & 1 & 4 & 16 & 64 & 256 & 1024 \\ 
     \midrule
    PSNR & 28.99 & 28.79 & 17.95& 22.81& 26.25&  27.93& 28.50& 28.66 \\
    SSIM& 0.869 & 0.870 & 0.285& 0.512& 0.714& 0.819& 0.856& 0.867 \\
    LPIPS & 0.185 & 0.181 & 0.611& 0.493& 0.351& 0.235& 0.178& 0.168 \\
    \bottomrule
    \end{tabular}}
    \vspace{-.5em}
    \caption{\textbf{Novel view synthesis --} quantitative results of different methods on MipNerf360 scenes. \methodname is finetuned on \stopthepop-trained scenes using 128 SPP, and rendered using varying number of samples. 
    }
\label{tab:forward-only-psnr-avg}
    \vspace{-1em}
\end{table}

%% file: tables/forward_time_eval.tex
\begin{table}
    \centering
    \resizebox{\linewidth}{!}{
    \setlength{\tabcolsep}{4pt}
     \begin{tabular}{@{}lccccccccc@{}} \toprule
     &  &  &  &\multicolumn{5}{c}{Ours-OpenGL (SPP)} \\ 
     & 3DGS-CUDA & 3DGS-OpenGL & \stopthepop &  1 & 2 & 4 & 8 & 16  \\ 
     \midrule
     T1000  & 65.08 & 100.28 & 75.91 & 16.23 & 18.95 & 24.71 & 37.51 & 61.25  \\
     RTX3090 & 8.14  & 32.15 & 9.48 & 3.25 & 4.18 & 6.42 & 15.31 & 18.48  \\
     RTX4090 & 5.60 & 20.70 & - & 1.85 & 2.05 & 2.86 & 6.71 & 8.00 \\
    \bottomrule
    \end{tabular}
    }
    \vspace{-.5em}
    \caption{\textbf{Forward pass time --} in milliseconds for different techniques on MipNerf360-trained scenes on three different GPUs. The values are averaged over all scenes and all views.} %
    \label{tab:forward-only-fps-avg}
    \vspace{-1em}
\end{table}

%% file: tables/training_eval.tex
\begin{table}
    \centering
    \resizebox{\linewidth}{!}{
    \setlength{\tabcolsep}{24pt}
     \begin{tabular}{@{}lcccc@{}} 
     \toprule
     &  & \multicolumn{3}{c}{Ours (SPP)} \\
     3DGS & \stopthepop & 1 & 8 & 128\\ 
     \midrule
     23.06 & 25.77 & 18.10 & 33.19 & 478.93\\
    \bottomrule
    \end{tabular}   
    }
    \vspace{-.5em}
    \caption{\textbf{Backward pass time -- } in milliseconds of the CUDA code for \methodname (as we vary SPP), 3DGS~\cite{Kerbl2023tdgs}, \stopthepop~\cite{stopthepop}, averaged over test views of the \mipnerfdataset scenes on an RTX 3090 GPU.}
    \label{tab:backwards_eval}
    \vspace{-1em}
\end{table}

%% file: figs/downsample/item.tex
\begin{figure}
  \vspace{-1em}
  \centering
  \includegraphics[width=.9\linewidth]{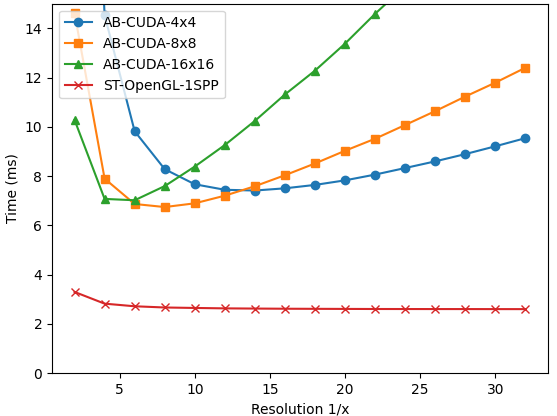}
  \vspace{-.5em}
  \caption{\textbf{Timings at lower resolution} -- Render timings (milliseconds) at different downsampled resolutions for our method (OpenGL) and alpha blending (CUDA) with different tile sizes on an NVIDIA RTX 3090.} 
  \vspace{-.5em}
  \label{fig:downsample}
\end{figure}

%% file: figs/taa/item.tex
\begin{figure}
\centering
\includegraphics[width=0.46\linewidth]{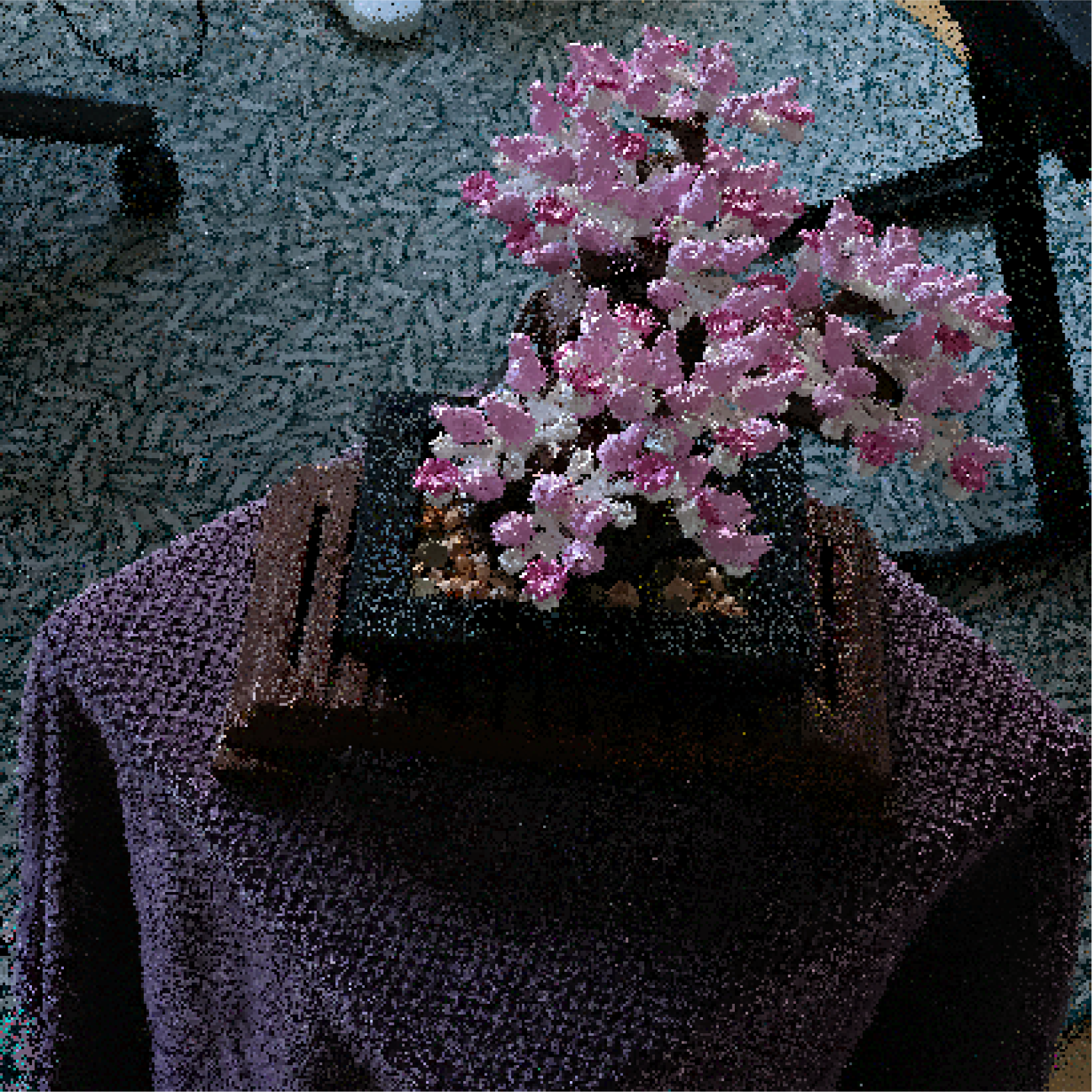}
\includegraphics[width=0.46\linewidth]{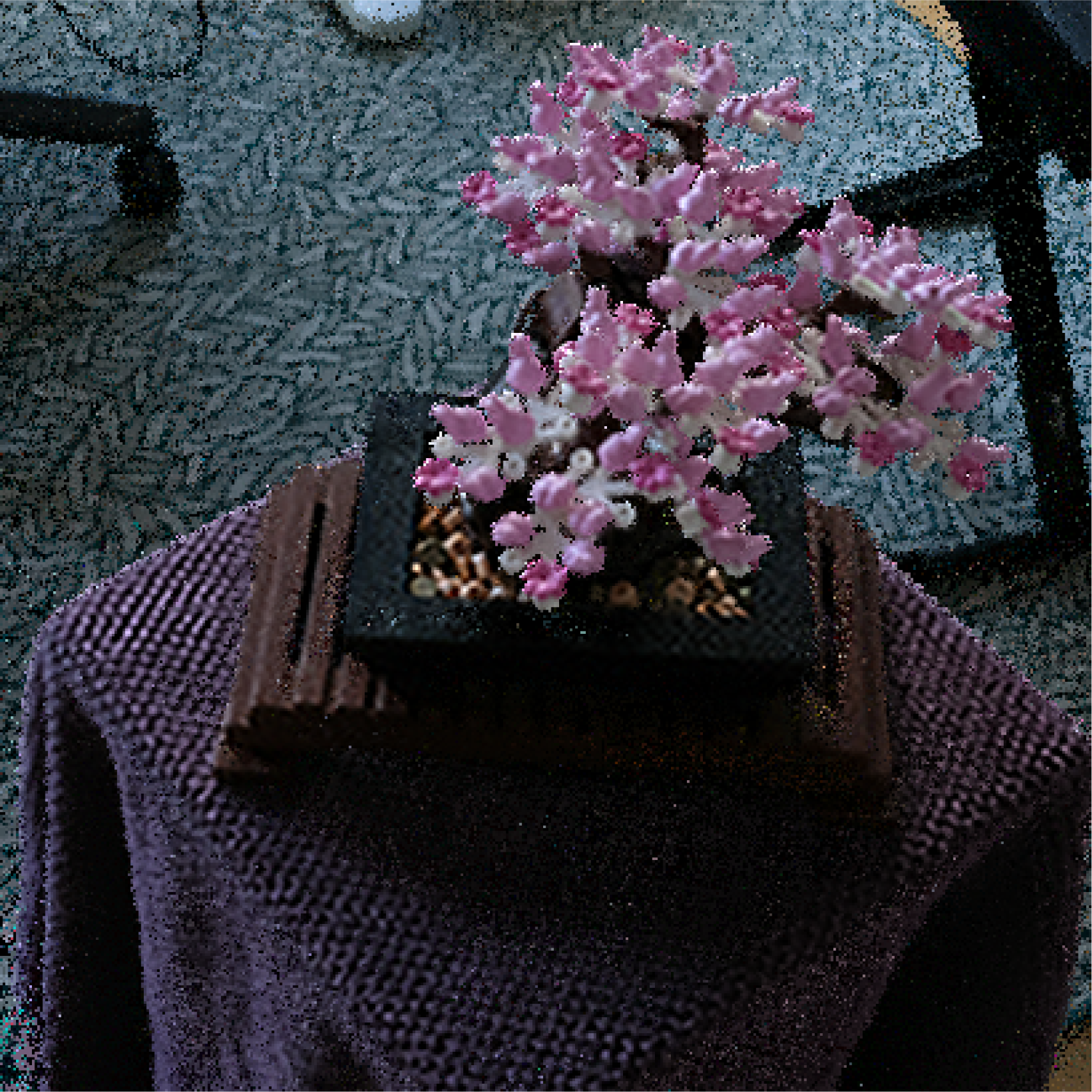}
\vspace{-.5em}
\caption{\textbf{Temporal anti-aliasing} -- \methodname at 1 SPP without (left) and with (right) TAA after moving along a trajectory for 70 frames.
TAA introduces a negligible latency increase. 
Video examples can be found in the supplementary.
\vspace{-1em}
}
\label{fig:taa_warping}
\end{figure}

%% file: figs/segmentation/seg.tex
\def \figtenw {0.48}
\begin{figure}
    \centering
    \begin{subfigure}{\figtenw\linewidth}
        \centering
        \includegraphics[width=\linewidth]{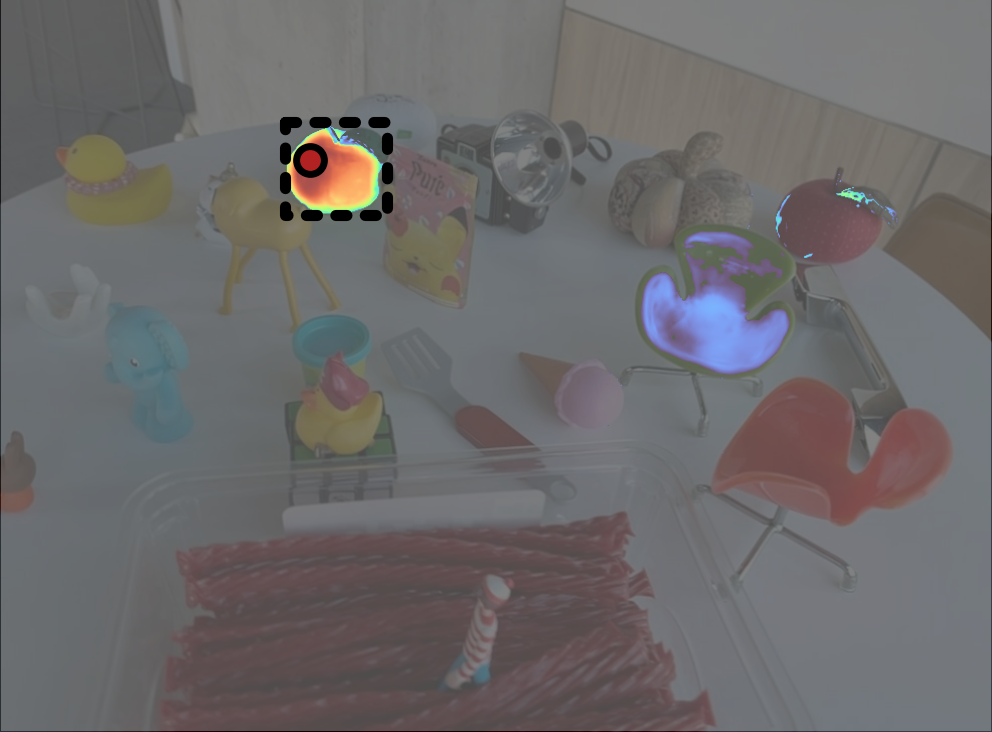}
        \caption{Green apple (AB)}
    \end{subfigure}
    \begin{subfigure}{\figtenw\linewidth}
        \centering
        \includegraphics[width=\linewidth]{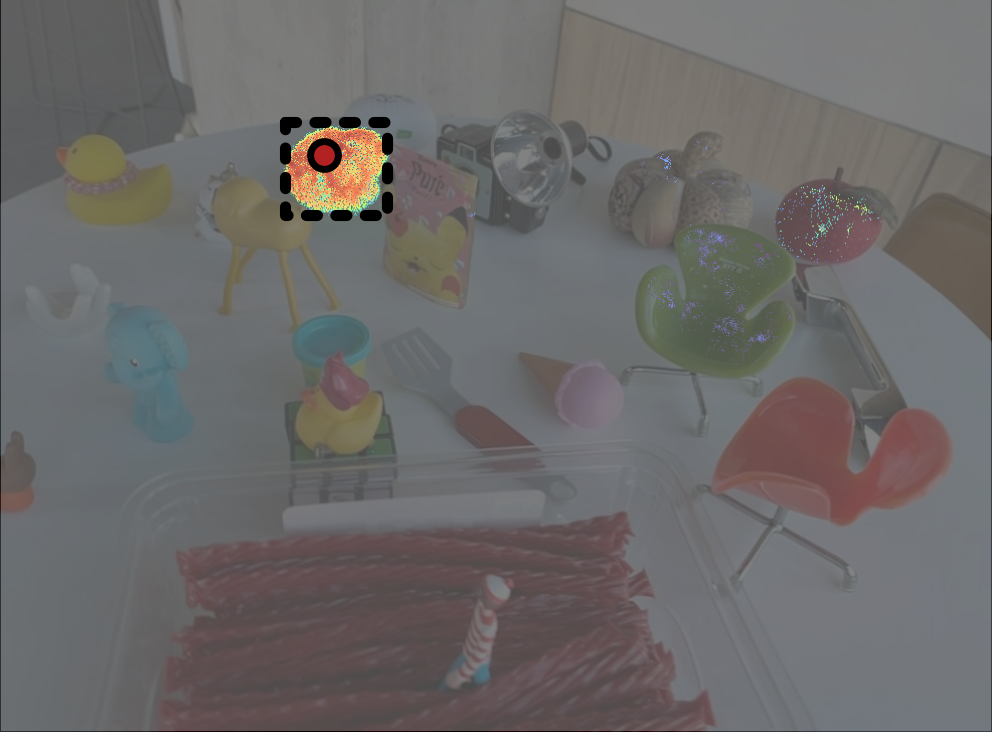}
        \caption{Green apple (ours)}
    \end{subfigure} 
    
    \begin{subfigure}{\figtenw\linewidth}
        \centering
        \includegraphics[width=\linewidth]{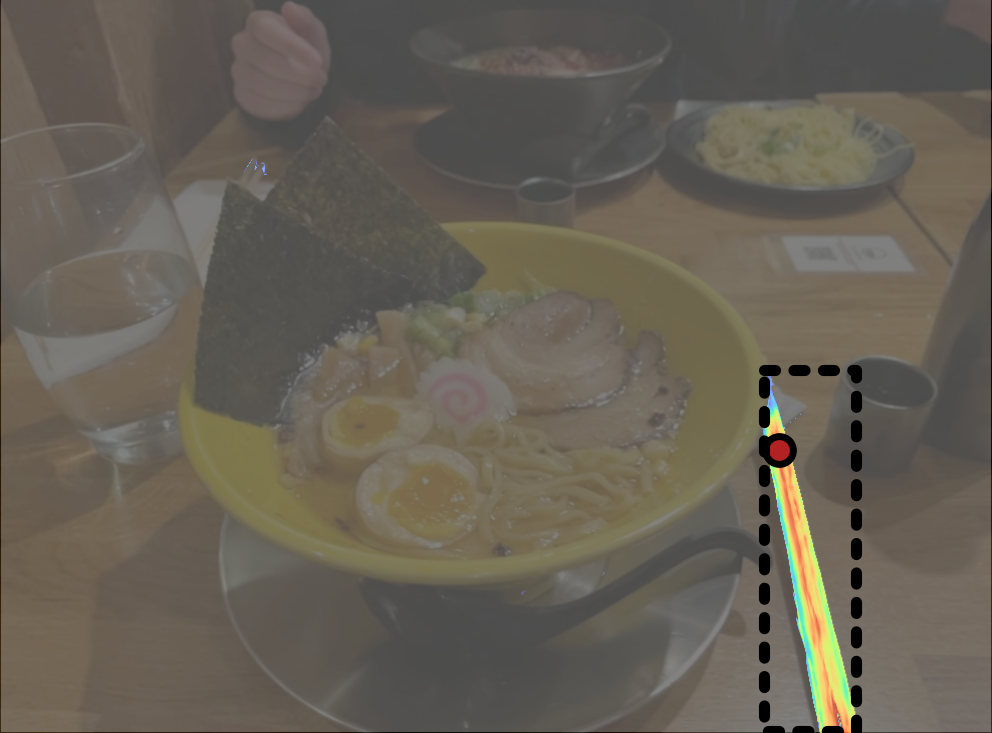}
        \caption{Chopsticks (AB)}
    \end{subfigure}
    \begin{subfigure}{\figtenw\linewidth}
        \centering
        \includegraphics[width=\linewidth]{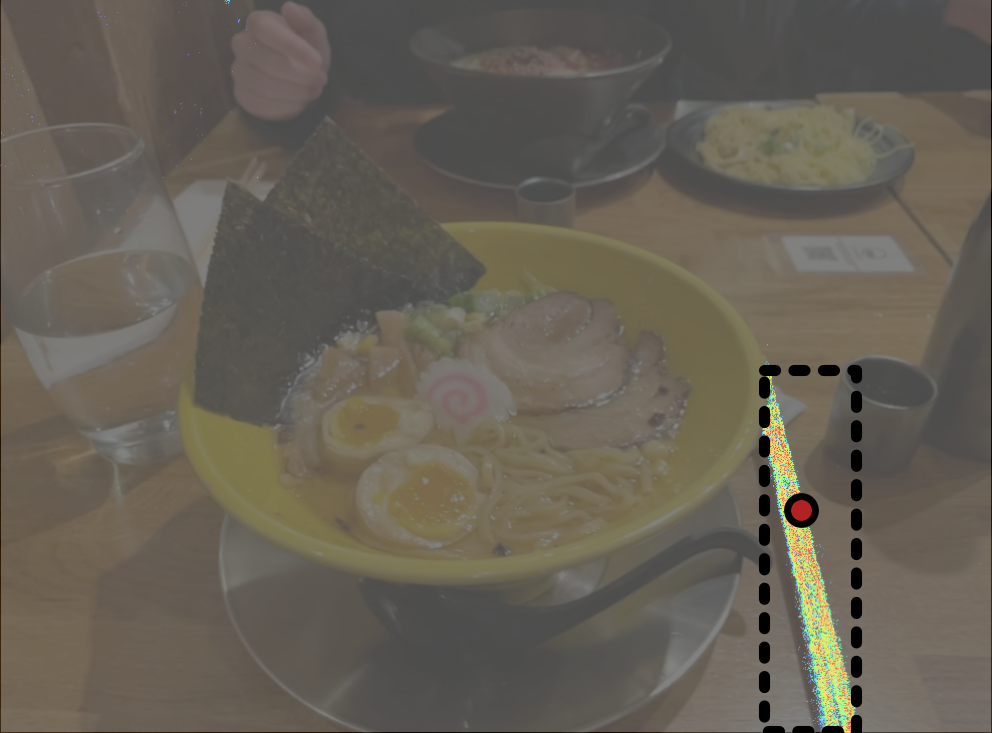}
        \caption{Chopsticks (ours)}
    \end{subfigure}
    \vspace{-.5em}
    \caption{\textbf{Open vocabulary localization -- }
      Qualitative comparison of open-vocabulary 3D object localization on the LERF dataset using alpha blending vs. \methodname at 1 SPP. 
    }
    \vspace{-1em}
    \label{fig:segmentation}
\end{figure}

%% file: tables/segmentation_eval.tex
\begin{table}[t]
    \centering
    \setlength{\tabcolsep}{12pt} %
    \resizebox{\linewidth}{!}{
    \begin{tabular}{@{}lccccc@{}}
    \toprule
              & Teatime & Ramen & Waldo-Kitchen & Figurines & \textbf{Mean} \\
    \midrule
    AB        & 88.1    & 73.2 & 95.5          & 80.4      & 84.3    \\
    Ours   & 88.1    & 73.2 & 86.4          & 80.4      & 82.0    \\
    \bottomrule
    \end{tabular}
    }
    \vspace{-.5em}
    \caption{\textbf{Open vocabulary localization -- } accuracy on the LERF dataset using alpha blending (AB) and \methodname at 1 SPP.
    }
    \label{tab:localization_eval}    
    \vspace{-1em}
\end{table}

%% file: sec/5_conclusions.tex
\section{Conclusion}
We proposed a sorting-free stochastic rendering method for 3D Gaussian splatting, that is, technically, compatible with any semi-transparent representation.
We derive both the forward (rendering) and backward (gradient) passes for stochastic rendering, allowing a quality vs. cost trade off, and renders without popping artifacts.
Our OpenGL implementation renders $2\times$~--~$4\times$ faster with reasonable quality, demonstrating high portability.

\paragraph{Limitations and future works}
While we obtain reasonable quality when rendering with low sample counts, our results are noisy. 
Future work could focus on reducing the variance of both forward and backward estimators, for example using importance sampling, stratified sampling~\cite{laine2011stratified} or control variates. 
Moreover, quality could further be improved by leveraging lightweight (neural) denoising algorithms, similar to how diffusion models are used in conjunction with rough 3D renderings~\cite{wu2025difix3d,ren2025gen3c}.

%% file: sec/X_suppl.tex
\clearpage
\setcounter{page}{1}
\setcounter{section}{0}
\maketitlesupplementary

\section{Per-scene results}

\Cref{tab:render_time_3090} and \Cref{tab:render_time_4090} show the times for different scenes and different rendering techniques.

\section{Tighter image-space bounding box}
\label{sec:gaussian_clipping}
As our method requires rasterizing all Gaussians without an early termination transmittance threshold, any reduction in the number of Gaussians within each tile helps in our CUDA implementation. 
\quad
3DGS projects a 3D Gaussian onto the image plane, resulting in a 2D Gaussian from which an axis-aligned bounding box around its center is computed in screen space. 3DGS uses a fix $\sigma{=}3$ to cutoff the Gaussian contribution.
We follow \citet{stopthepop} approach and use $t_O = \sqrt{2 \log ({\alpha}/{\epsilon_O} )}$ with $\epsilon_o = {1}/{255}$ in all our results, including 3DGS. 

Given a 2D screen-space Gaussian, 3DGS computes the major and minor eigenvectors and their corresponding eigenvalues, \(\lambda_1\) and \(\lambda_2\). The larger eigenvalue provides a bound on the Gaussian’s radius in screen space, scaled by $t_O$. The radius is then used to form an axis-aligned \emph{square} bounding box around the center of the projected
Gaussian.

However, for elongated Gaussians, this square approximation introduces unnecessary overhead in a tile-based implementation, increasing the number of per-tile-Gaussians. Instead, a tighter bounding box can be derived by accounting for both eigenvalues
and their associated eigenvectors. Specifically, in each eigenvector direction
\(\mathbf{v}_i\), we set
\begin{align}
  \Delta_i \;=\; \sqrt{t_O\,\lambda_i}, \quad i=1,2,
\end{align}
In our OpenGL implementation, we also compute the bounding-box corners as described above, which yields a tighter fit, without requiring them to be axis-aligned. Moreover, by increasing the number of corner samples, we can further refine this bounding region with minimal additional effort. 

\section{Temporal anti-aliasing}
At the first frame, we record each pixel’s world-space position \(\worldpos\) and color. For each subsequent frame, these 3D points are projected into the new view to produce a warped version of the previous average frame’s image. We then compare the warped 3D coordinates with those computed for the current frame. If they closely match, we blend both the color and the \(\worldpos\) values. Otherwise, the sample count for that pixel is reset.
Let \(c_t\) denote the color image at time \(t\), and \(\mu_{t-1}\) be the accumulated warped average from previous frames. We update the color in the current frame by:
\begin{equation}
    \bar{c}_t 
    = \frac{N_{t-1}}{N_{t-1} + 1} \mu_{t-1} 
    + \frac{1}{N_{t-1} + 1} c_t,
\end{equation}
where \(N_{t-1}\) represents the number of accumulated samples up to frame \(t-1\). The same operation applies to the 3D coordinates.
\quad
We note this average position is well defined for surfaces, but poorly approximated in less confident ``volumetric'' regions.
We define new samples that are further from the warped mesh by more than a threshold $\thresh$ to be occluded, and reset accumulation for those pixels.

\input{tables/rendering_eval_details}

\section{Free-flight distance sampling}
For completeness, we describe the free-flight distance sampling we use for volumetric intermixing in detail. Following~\cite{condor2024}, we define a volumetric density for a 3D Gaussian:
\begin{align}
     \sigma(\vx) = \sigma_t \exp\left(-\frac{1}{2} (\vx - \mu)^T \Sigma^{-1} (\vx - \mu)\right),
\end{align}
where $\sigma_t$ scales the density of the entire Gaussian uniformly and $\vx$ is a 3D position. The integral of this density along a ray $(\vo, \vd)$ can be analytically computed:
\begin{align}
\int_0^t \sigma(\vx_t) \diff t = & \sigma_t \frac{\sqrt{\pi}}{\sqrt{2}c}  \exp\left((a^2 - b)/2\right) \nonumber \\
&\cdot \left(\erf\left((a + tc)/\sqrt{2}\right) - \erf\left(a/\sqrt{2}\right)  \right) 
\end{align}
where $t$ is the distance along the ray and we define:
\begin{align}
a &= \frac{(\vo - \mu)^T \Sigma^{-1} \vd}{c} & 
b &= (\vo - \mu)^T \Sigma^{-1} (\vo - \mu) \nonumber \\ 
c &= \sqrt{\vd^T \Sigma^{-1} \vd}. \nonumber
\end{align}%

In turn, this enables the analytical computation of the free-flight PDF $p(t) = \sigma(\vx_t) \exp(-\int_0^t \sigma(\vx_s) \diff s)$ and its CDF, which is simply $1 - \exp(-\int_0^t \sigma(\vx_s) \diff s)$. This allows to derive an analytical inverse CDF sampling routine~\cite{condor2024}:
\begin{align*}
d &= \erf\left( \sqrt{2} a \right) - \sqrt{\frac{2}{\pi}} c \sigma_t^{-1} \exp\left((b-a^2)/2\right) \log(1 - u) \\
t &= \begin{cases}
    \frac{1}{c} \left(-a + \erf^{-1}(d)/d \right) &\text{if } d \in (-1, 1) \\ 
    \infty &\text{otherwise},
\end{cases}
\end{align*}
where $u \sim \mathcal{U}(0,1)$. The distance $t$ computed using this routine is distributed proportional to $p(t)$.

%% file: tables/rendering_eval_details.tex
\begin{table}[t]
    \centering
    \small
    \resizebox{\linewidth}{!}{
    \begin{tabular}{lccccccc}
        \toprule
        & ST@1 & ST@2 & ST@4 & ST@8 & ST@16 & AB-GL & AB-CUDA \\
        \midrule
        Room    & 1.91  & 3.14  & 5.70  & 16.24  & 20.49  & 13.73  & 5.92  \\
        Bonsai  & 1.55  & 2.47  & 4.43  & 12.58  & 15.64  & 10.13  & 4.50  \\
        Counter & 2.08  & 3.62  & 6.70  & 19.39  & 24.10  & 12.24  & 5.75  \\
        Kitchen & 2.83  & 4.51  & 7.96  & 22.33  & 27.28  & 17.86  & 7.41  \\
        Stump   & 4.12  & 4.42  & 5.09  & 8.98   & 10.72  & 35.80  & 8.90  \\
        Bicycle & 5.13  & 5.48  & 6.27  & 11.89  & 14.00  & 64.14  & 12.44 \\
        Garden  & 5.13  & 5.62  & 6.83  & 13.77  & 16.11  & 73.19  & 12.07 \\
        \midrule
        \textbf{Average} & 3.25  & 4.18  & 6.42  & 15.31  & 18.48  & 32.16  & 8.14  \\
        \bottomrule
    \end{tabular}}
    \caption{Average rendering time on 3090 GPU}
    \label{tab:render_time_3090}
\end{table}

\begin{table}[t]
    \centering
    \small
    \resizebox{\linewidth}{!}{
    \begin{tabular}{lccccccc}
        \toprule
        & ST@1 & ST@2 & ST@4 & ST@8 & ST@16 & AB-GL & AB-CUDA \\
        \midrule
        Room    & 0.99  & 1.40  & 2.33  & 6.98  & 8.35  & 8.87  & 3.76  \\
        Bonsai  & 0.83  & 1.16  & 1.89  & 5.40  & 6.39  & 6.42  & 2.89  \\
        Counter & 1.03  & 1.59  & 2.75  & 8.33  & 9.84  & 8.20  & 3.83  \\
        Kitchen & 1.42  & 2.05  & 3.41  & 9.60  & 11.20 & 11.36 & 4.89  \\
        Stump   & 2.15  & 2.29  & 2.64  & 4.26  & 5.09  & 24.15 & 6.32  \\
        Bicycle & 2.71  & 2.88  & 3.35  & 5.63  & 6.50  & 41.17 & 9.08  \\
        Garden  & 2.79  & 3.01  & 3.66  & 6.78  & 7.62  & 44.74 & 8.46  \\
        \midrule
        \textbf{Average} & 1.85  & 2.05  & 2.86  & 6.71  & 8.00  & 20.70 & 5.60  \\
        \bottomrule
    \end{tabular}}
    \caption{Average rendering time on 4090 GPU}
    \label{tab:render_time_4090}
\end{table}